\documentclass[10pt,twocolumn,letterpaper]{article}

\usepackage{cvpr}              
\usepackage{float}
\usepackage{algorithm}
\usepackage{algorithmic}
\usepackage{caption}
\usepackage{dsfont}
\usepackage{tabularx}
\usepackage{diagbox}
\usepackage[accsupp]{axessibility}

\usepackage[dvipsnames]{xcolor}

\newcommand{\methodnameacr}{MASH\xspace}
\newcommand{\methodname}{MAsked and SHuffled Blind Spot Denoising\xspace}

\definecolor{cvprblue}{rgb}{0.21,0.49,0.74}
\usepackage[pagebackref,breaklinks,colorlinks,citecolor=cvprblue]{hyperref}

\def\paperID{15942} 
\def\confName{CVPR}
\def\confYear{2024}

\title{Masked and Shuffled Blind Spot Denoising for Real-World Images}

\author{Hamadi Chihaoui\\
Computer Vision Group\\
University of Bern, Switzerland\\
{\tt\small hamadi.chihaoui@unibe.ch}
 \and
Paolo Favaro\\
Computer Vision Group\\
University of Bern, Switzerland\\
{\tt\small paolo.favaro@unibe.ch}
}

\begin{document}
\maketitle


\begin{abstract}
We introduce a novel approach to single image denoising based on the Blind Spot Denoising principle, which we call \textit{\methodname} (\textit{\methodnameacr}). We focus on the case of correlated noise, which often plagues real images. \textit{\methodnameacr} is the result of a careful analysis to determine the relationships between the level of blindness (masking) of the input and the (unknown) noise correlation. Moreover, we introduce a shuffling technique to weaken the local correlation of noise, which in turn yields an additional denoising performance improvement. We evaluate \textit{\methodnameacr} via extensive experiments on real-world noisy image datasets. We demonstrate state-of-the-art results compared to existing self-supervised denoising methods. Website:\\\url{https://hamadichihaoui.github.io/mash}.
\end{abstract}

\section{Introduction}



The removal of noise from real images, \ie, image denoising, is a fundamental and still open problem in image processing despite having a long history of dedicated research (see \cite{gu2019brief} for an overview of the classic and recent methods). In classic methods, the primary strategies involve manually designing image priors and optimization techniques to enhance both reconstruction accuracy and speed. In contrast, in the context of deep learning methods, neural networks naturally introduce a very powerful prior for images \cite{ulyanov2018dip} and provide models that could perform denoising efficiently at inference time. These innate capabilities of neural networks opened the doors to a wide range of methods that could not only learn to denoise image from examples of noisy and clean image pairs, but, even more remarkably, directly from single noisy images \cite{krull2019n2v,xie2020n2s, huang2021neighbor2neighbor,wang2022blind2unblind}.

In this work, we push the limits of these advanced methods one step further. We focus on the family of methods called Blind Spot Denoising (BSD)\cite{krull2019n2v}, since it provides a powerful and general framework. Moreover, we consider the case where only a single image is used for denoising (\ie, we do not rely on a supporting dataset). As also observed by Wang et al~\cite{wang2022blind2unblind}, training on a dataset may not generalize well on new data, where the noise distribution is unknown. This is particularly true for real images, where noise is often correlated. In these settings, most modern methods find it challenging to handle non-iid data. 

In our work, similar to the approach in \cite{quan2020self2self}, we explore the more general setting of random masking beyond the single blind spot method introduced in \cite{krull2019n2v}.
In our analysis, we uncover valuable connections between the performance of Blind Spot Denoising (BSD) methods trained with various input masking techniques and the degree of noise correlation. Surprisingly, we observe that models trained with a higher masking ratio tend to perform better when dealing with highly correlated noise, whereas models trained with a lower masking ratio excel in denoising tasks with iid noise. This discovery offers two key contributions: 1) it provides a method to estimate the unknown level of noise correlation, and 2) it offers a strategy for achieving enhanced denoising performance. Furthermore, our analysis reveals that noise correlation significantly hampers the denoising capabilities of BSD models. This suggests that a more radical approach would be to directly eliminate the correlation in the input data. An intuitive method to achieve this would involve randomly permuting all pixels that correspond to the same clean-image color intensity. However, this presents a classic chicken and egg dilemma, as we would typically need the clean image to perform the permutation, yet the clean image is precisely what we are trying to restore. To tackle this challenge, we utilize an intermediate denoised image as a pseudo-clean image to define the permutation set. Furthermore, given that adjacent pixels are likely to have similar color intensities, we focus on shuffling only pixels within small neighborhoods. We incorporate these insights into a novel method called \methodname (\methodnameacr{}), which we elaborate on further in \cref{sec:mash}.
Our contributions are summarized as follows
\begin{itemize}
    \item We provide an analysis of BSD, showcasing the impact of various masking ratios on correlated noise and presenting a method for estimating the noise correlation level;
    
    \item We introduce \methodnameacr{}, an enhanced version of BSD that dynamically selects the optimal masking ratio. We also introduce the local pixel shuffling technique to address noise correlation at its source;

    \item \methodnameacr{} demonstrates marked enhancements over the baseline BSD and attains on par or better results in real-world denoising across multiple datasets.
\end{itemize}


\section{Related Work}
In this section, we only focus on work that is closely related to unsupervised image denoising.

\noindent\textbf{Non-learning-based image denoisers.} 
Classic image denoisers \cite{burger2012image, zhang2010two, talebi2013global} manually define image priors of what a clean image is. Some approaches explore sparse representations \cite{bao20130sparsecoding, elad2006sparsecoding}, while others benefit from the patch recurrence prior \cite{zontak2013separating}.
BM3D~\cite{dabov2009bm3d}, which applies collaborative filtering to similar patches, is one of the best-known non-local methods, due to its high performance across several benchmarks. NLM~\cite{buades2011nlm} and WNNM~\cite{gu2011nlm} also relate similar patches and use them to reduce noise through a form of implicit averaging. 

\noindent\textbf{Unsupervised learning-based image denoisers.}
The unsupervised learning-based approaches can be classified into two categories based on their training data. The first category is the dataset-based one, which uses a dataset of noisy images to train a denoising model. 
The second category is the single-image one, which learns a denoiser from a single image at a time.\\
1) \textit{Dataset-based learning approaches.}\\
Noise2Noise~\cite{lehtinen2018n2n} shows that it is possible to train a denoising neural network without the need for clean images. Indeed, using only pairs of noisy images of the same scene as input/output training data leads to a comparable performance of standard noisy/clean image pair supervised training. Some approaches \cite{krull2019n2v,xie2020n2s, huang2021neighbor2neighbor,wang2022blind2unblind} go one step further and remove the need for noisy image pairs altogether.
The main contributions in these approaches are ways to avoid learning the trivial solution of the identity mapping. 
 Recently, \cite{zhou2020awgn} proposed the use of pixel-shuffle downsampling (PD) to weaken the spatial correlation of structured noise. In a similar vein, AP-BSN \cite{lee2022ap} utilized two PDs with different strides for training and testing. CVF-SID \cite{neshatavar2022cvf} attempted to separate the latent image and noise from the noisy input through a cyclic module and self-supervised losses. LUD-VAE \cite{zheng2022learn} established two hidden variables for the noise domain and the latent image domain and optimize for them using unpaired data. \cite{jang2021c2n, kousha2022modeling} explicitly learned a model for real-world noise. These methods rely on noisy or unpaired data, and generalization issues persist.\\
2) \textit{Test-time training approaches.} \\
A learning-based image denoising approach without any a-priori dependence on training samples is the least demanding option to denoise a given noisy image. 
There are a few methods that use a single noisy image to train a deep neural network for denoising. The DIP~\cite{ulyanov2018dip} shows that the inductive bias of convolutional neural networks (CNN) favors the learning of noise-free texture patterns rather than noisy ones, when trying to reconstruct a degraded image. The method trains a CNN to generate a given degraded image from a random input, while early stopping is used as a regularization. DIP is highly sensitive to the choice of the stopping time and thus is not practical as a denoising method. 
ScoreDVI \cite{cheng2023score} recently proposes a method for denoising a single image by exploiting score priors embedded in MMSE. 
\cite{zheng2020unsupervised} addresses the complexity of real noise by mapping the noisy image into a latent space in which the additive white Gaussian noise (AWGN) assumption holds. An encoder-decoder is used for the mapping and an off-the-shelf Gaussian denoiser is used to denoise the encoded image. Finally the image is decoded back to the original space. \cite{quan2020self2self} is a self-supervised approach that use pairs of Bernoulli-sampled instances of the input noisy image to train a denoiser with dropout. The denoiser predicts the masked pixels based on the visible ones. The final denoised image is the average of the predictions generated from multiple instances of the trained model with dropout. 

\methodnameacr{} belongs to the category of BSD methods applied to a single noisy image. However, in contrast to the above methods, it uses an optimal masking ratio that is automatically estimated and introduced the technique of local pixel shuffling. To the best of our knowledge, these contributions have not been presented before.







\section{Unsupervised Single Image Denoising}
\label{sec:mash}
In this section, we begin by revisiting the concept of BSD as introduced in a prior work by Krull et al. \cite{krull2019n2v}. Subsequently, we conduct experimental analyses to delve deeper into understanding the effects of various design factors in the BSD method on its performance. For this purpose, we generate a synthetic noisy dataset with noise ranging from independent and identically distributed (iid) to highly correlated, using which we train a base blind spot network with differing levels of "blindness" (i.e., image masking ratios). These empirical observations lead us to create a novel self-supervised blind-spot framework for unsupervised single image denoising, referred to as \methodname (\methodnameacr{}), which achieves state-of-the-art performance.

\begin{figure}[t]

\subfloat[\centering $\beta=0$]{\includegraphics[width=0.155\textwidth]{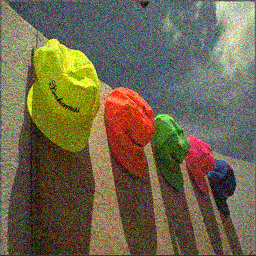}}
\hfill
\subfloat[\centering $\beta=0.5$
]{\includegraphics[width=0.155\textwidth]{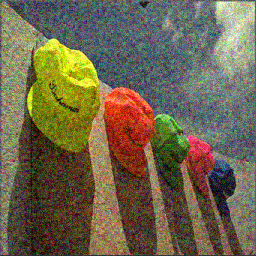}}
\hfill
\subfloat[\centering $\beta=1$]{\includegraphics[width=0.155\textwidth]{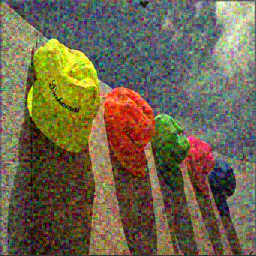}}
\hfill
    \caption{\label{fig:generated_noisy_images} Samples of generated noisy images depending on the spatial correlation level. From left to right: noisy image with iid noise, noisy image with moderately correlated noise and noisy image with heavily correlated noise}
\end{figure}

\subsection{Revisiting Blind Spot Denoising (BSD)}
BSD operates as a self-supervised technique, meaning that its training process does not necessitate pairs of noisy images or pairs of noisy and clean images. Instead, only noisy images are employed. The method applies a masking scheme that hides part of the image at the input and then aims to predict the same hidden part at the output. In this paper, we apply this idea to a single image at a time.
Formally, considering the noisy observation $\mathbf {y}\in \mathbb{R}^{H\times W\times C}$ corresponding to the clean image $\mathbf {x}\in \mathbb{R}^{H\times W\times C}$, where $H\times W$ represents the image dimensions and $C$ denotes the color channels (usually $1$ for grayscale or $3$ for RGB), the objective of BSD is to minimize the following empirical risk:
\begin{equation}
    \mathop {\arg \min\limits _{\theta } } \sum_{\mathbf{i}\in \Omega} \Vert f_{\theta }(\mathbf {y}_{\text{masked}(\mathbf{i})})[\mathbf{i}]-\mathbf {y}[\mathbf{i}]\Vert _2^2
    \label{eq:BSD}
\end{equation}
In \cref{eq:BSD}, $f_\theta:\mathbb{R}^{H\times W\times C}\rightarrow \mathbb{R}^{H\times W\times C}$ represents the denoising model realized through neural networks with parameters $\theta$, which transforms noisy images into denoised versions. Here, $\mathbf{y}_{\text{masked}(\mathbf{i})}$ denotes the image $\mathbf{y}$ where the pixel $\mathbf{i}\in \Omega$ ($\Omega=[1,\dots,H]\times [1,\dots,W]$), has been masked. Additionally, $\mathbf{y}[\mathbf{i}]$ refers to the RGB color at the pixel $\mathbf{i}$.
Notice that the above formulation is used for a single image $\mathbf{y}$. It is also possible to extend it to a dataset of images simply by taking the expectation of the above loss with respect to the distribution of $\mathbf{y}$. However, as already mentioned, in this paper we only consider denoising a single image without additional data.
In our terminology, we also refer to \emph{signal} when talking about the clean image. BSD assumes that noise at each pixel is statistically independent of its neighbors (thus a noise instance cannot be predicted by nearby noise), while a pixel of the signal (\ie, the clean image) is spatially correlated to its neighboring pixels. These assumptions are key in enabling the separation of noise from signal. 
More in general, the BSD method can also be seen as an extreme case of masking or sparse image reconstruction/inpainting such as MAE \cite{he2022masked}, where the masking consists of a single pixel. 
In practice, the BSD method avoids trivial solutions, such as the identity mapping (simply reconstructing the noisy image), only as long as $f_\theta$ does not overfit the data. 
This is particularly important in the context of single image denoising, where the training dataset is limited in size. Therefore, considering that the degree of masking may have a regularizing effect on $f_\theta$, we investigate the influence of this design choice on single image denoising.

\subsection{Diving Deep into Blind Spot Denoising}

In this section, our goal is to empirically evaluate the performance of the BSD method by examining the effects of two key factors: 1) the masking ratio used in the BSD method and 2) the characteristics of the noise (specifically, its level of spatial correlation). To conduct this analysis, we generate a denoising dataset synthetically using images from the Kodak dataset \cite{franzen1999kodak}. We model the noise using a multivariate normal distribution with a variance-covariance matrix $\Sigma$ to simulate real-world correlated noise. The correlation $\Sigma[\mathbf{i},\mathbf{j}]$ between pixels $\mathbf{i}$ and $\mathbf{j}$ in the set $\Omega$ is defined as follows:
 \begin{equation}
  \Sigma[\mathbf{i},\mathbf{j}] =
    \begin{cases}
      \sigma^2 & \text{ $\mathbf{i} = \mathbf{j}$}\\
      \beta \frac{k - \|\mathbf{i}-\mathbf{j}\|}{k} \sigma^2 & \text{$ 0 <  \|\mathbf{i}-\mathbf{j}\| \leq  k$}\\
      0 & \text{ otherwise}
    \end{cases}       
\end{equation}
 where $\|\mathbf{i}-\mathbf{j}\|$ is the distance between the pixels $\mathbf{i}$ and $\mathbf{j}$ and $k>0$ denotes the correlation kernel width. 
The parameter $\beta>0$ is used to control the level of spatial correlation in the noise. A higher value of $\beta$ indicates stronger correlation in the noise. We define three distinct regimes based on the noise correlation:
1) The \emph{iid regime}, where $\beta = 0$, and the noise is independent and identically distributed (iid);
2) The \emph{moderately correlated regime}, where $\beta = 0.5$;
3) The \emph{heavily correlated regime}, where $\beta = 1.0$;
We set $k = 3$ and $\sigma = 25$. An extension to this analysis, considering lower noise levels ($\sigma = 15$) and higher noise levels ($\sigma = 40$), is provided in the supplementary material due to space constraints. In this extension, we draw similar conclusions to those obtained for $\sigma = 25$. \cref{fig:generated_noisy_images} illustrates an example of an image with synthetically generated noise for each of the defined regimes.

We define the \emph{blindness mask} $\mathbf{m}:\Omega\mapsto \{0,1\}^{{H\times W\times C}}$,
where $\Omega$ represents the set of pixels in our noisy image $\mathbf{y}$. 
\begin{equation}
    \mathbf{m[\mathbf{i}]} = 
    \begin{cases}
      0 & \text{with probability $\tau$ };\\     
      1 & \text{with probability $1 - \tau$  }.
    \end{cases}   
     \label{eq:mask_selection}
\end{equation}
$\mathbf{m}$
is of the same size as the input images and is determined by a masking ratio parameter $\tau$. The masking ratio $\tau$ represents the proportion of zeros in the mask $\mathbf{m}$, calculated as the number of zeros divided by the total number of pixels in the image ($HWC$). 
We adopt a similar BSD formulation as \cite{quan2020self2self} with the blindness mask via the following loss

\begin{equation}
    L(\theta) = \mathbb{E}_\mathbf{m} \left[ \| (\mathbf{1} - \mathbf{m}) \odot   (f_{\theta}(\mathbf{y} \odot \mathbf{m}) - \mathbf{y}) \|_2^2 \right]
    \label{eq:loss_bsd}
\end{equation}
where $\mathbb{E}_\mathbf{m}$ denotes the expectation with respect to the mask $\mathbf{m}$ and $\odot$ denotes the element-wise product. 
\subsubsection{Impact of the masking ratio on BSD performance}
\begin{figure}[t]
 \setkeys{Gin}{width=1.0\linewidth}
    \captionsetup[subfigure]{skip=0.5ex,
                             belowskip=1ex,
                             labelformat=simple}
    \renewcommand\thesubfigure{}
\includegraphics[]{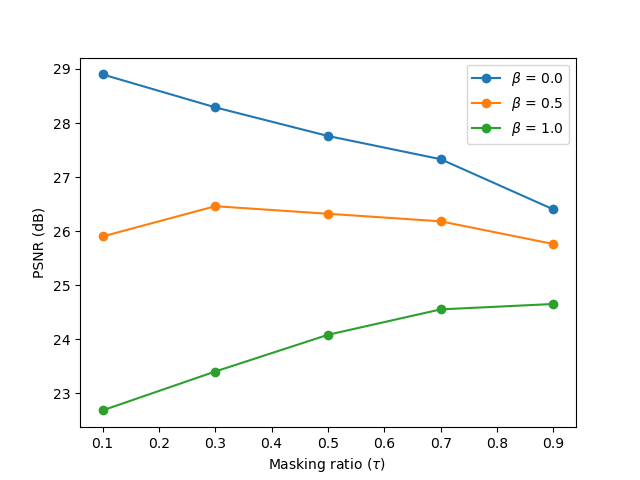}
\caption{\label{fig:denoising_performance_based_on_masking} Impact of the masking ratio $\tau$ on the generalized BSD denoising performance (PSNR). On the horizontal axis we consider several masking ratios $\tau$ and for each we train a BSD model on data with different levels of correlation $\beta$. The optimal performance of the trained model shows a strong correlation between the masking ratio an the noise correlation. Low masking benefits the training on data with iid noise and high masking benefits the training on data with highly correlated noise.}
\end{figure}

We investigate the impact of the masking ratio $\tau$ on the denoising performance depending on the correlation magnitude controlled by $\beta$. We train a denoising network by minimizing the loss ~\eqref{eq:loss_bsd} with different masking levels in the three noise regimes (\ie, with different correlation magnitudes $\beta$). 
The results are shown in \cref{fig:denoising_performance_based_on_masking}. We plot the Peak Signal-to-Noise Ratio (PSNR) values in decibels across the three regimes  for various masking configurations (displayed on the x-axis). We notice that the effectiveness of our generalized BSD is greatly influenced by: 1) the correlation of the noise, and 2) the masking ratio. Specifically, in the case of independent and identically distributed (iid) noise, we discovered that a lower masking ratio results in the best performance. Conversely, in scenarios where the noise exhibits high correlation, a higher masking ratio produces superior performance. In situations of moderate correlation, an intermediate masking ratio (ranging from 0.3 to 0.5) is deemed optimal. As observed in the analysis of DIP \cite{ulyanov2018dip} a neural network model, such as $f_\theta$, tends to first fit clean image patterns and then noise patterns. In scenarios of high spatial noise correlation, the noise tends to exhibit patterns resembling textures present in clean images (refer to the rightmost image in \cref{fig:generated_noisy_images}). Consequently, in cases of increased noise correlation, the efficacy of BSD could potentially improve with greater regularization, \ie, a higher masking ratio $\tau$.
\begin{figure}[t]
 \setkeys{Gin}{width=1.0\linewidth}
    \captionsetup[subfigure]{skip=0.5ex,
                             belowskip=1ex,
                             labelformat=simple}
    \renewcommand\thesubfigure{}
\includegraphics[]{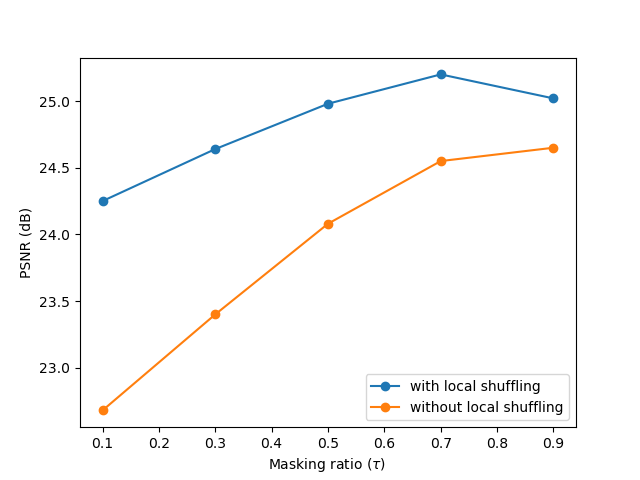}
\caption{\label{fig:denoising_performance_based_on_shuffling} Impact of the local pixel shuffling on the denoising performance (PSNR) when noise is highly correlated. The shuffling of image regions that are approximately constant destroys the noise correlation. This brings a consistent benefit across all masking ratios.}
\end{figure}
\subsubsection{Handling correlation with local pixel shuffling}
By comparing the top performances of BSD under each noise regime in \cref{fig:denoising_performance_based_on_masking}, we can readily observe a significant decrease in performance in the highly correlated noise scenario compared to the iid case, despite both scenarios having the same noise level. 
This observation prompts the consideration of another strategy to enhance performance in the presence of correlated noise.
The idea is to reduce noise correlation without altering the underlying image signal. One intuitive approach is to identify sets of noisy pixels corresponding to identical colors in the clean image and then randomly permute them. By doing so, we can disrupt the spatial correlation of these pixels without affecting the clean image structure. A practical method to implement this concept is to utilize the predicted denoised image as a pseudo-clean image. This pseudo-clean image can be leveraged to identify sets of iso-intensity pixels. Since neighboring pixels are more likely to share similar intensities, the swapping process can be performed locally to maintain coherence in the image.
To implement this idea, we first divide the image into two categories of regions: regions with nearly constant intensity and regions with texture. We define the image partition denoted as $\Omega_\text{const}$ as the set of pixels where the local neighborhoods, such as those defined by $4\times 4$ pixel blocks, exhibit similar or identical color intensity values. The remaining partition is  categorized with texture ( significant variation in intensity patterns).

\begin{figure}[ht]

\subfloat[\centering ]{\includegraphics[width=0.155\textwidth]{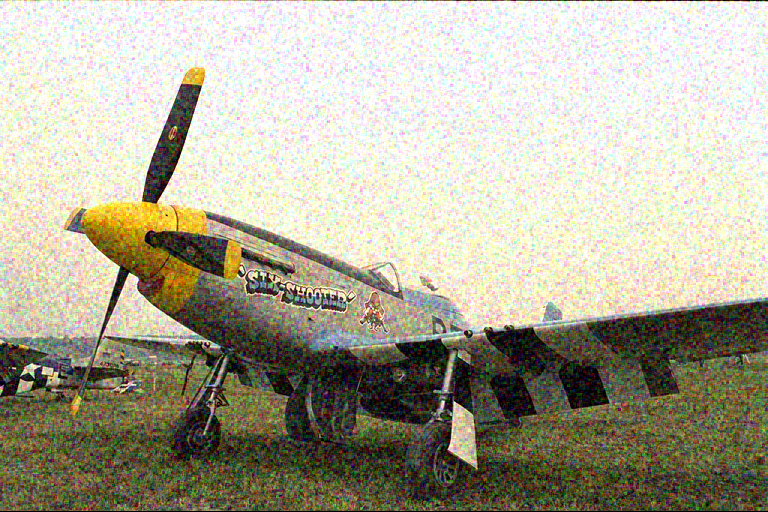}}
\subfloat[\centering 
]{\includegraphics[width=0.155\textwidth]{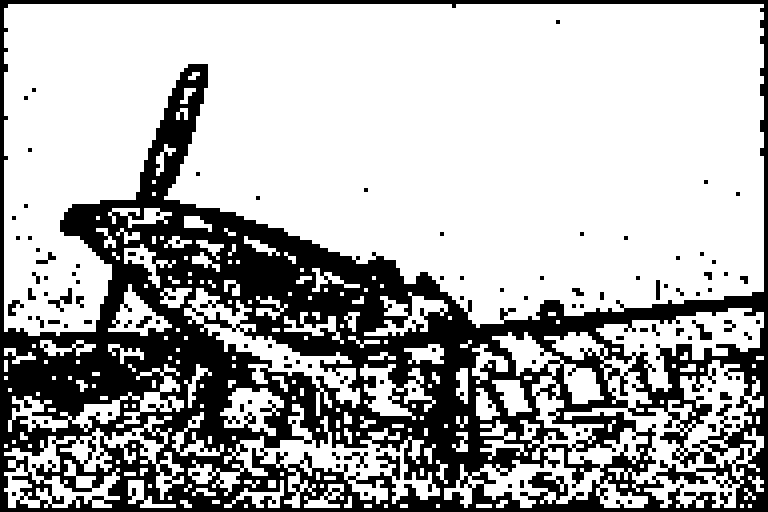}}
\subfloat[\centering ]{\includegraphics[width=0.155\textwidth]{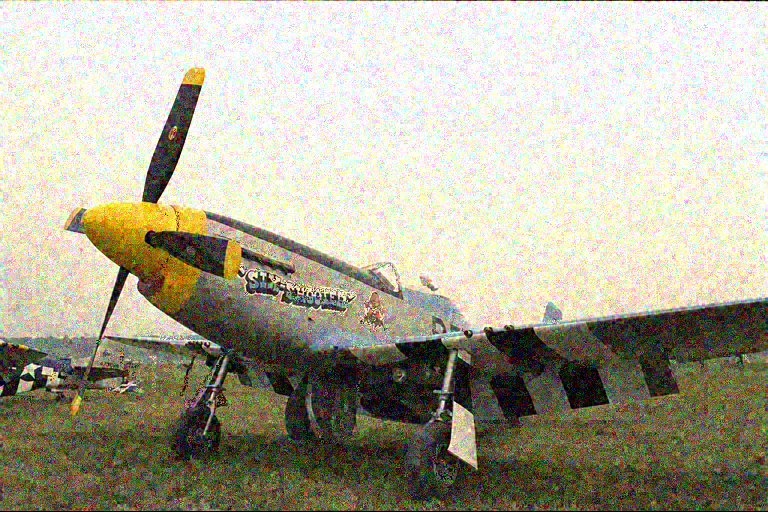}}
    \caption{\label{fig:flattness_images} (a) Original noisy image (b) Mask capturing region flatness derived from pseudo-clean prediction (c) Noisy image with local pixel permutation on flat regions.}
\end{figure}

We define $\mathbf{c}(\mathbf{y})$ as the mapping that assigns a value of $1$ to the pixels within the constant intensity regions ($\Omega_\text{const}$) of $\mathbf{y}$ and a value of $0$ to all other pixels within the image domain $\Omega$. $\mathbf{c(y)}$ helps to differentiate between the constant intensity regions and the textured regions within $\mathbf{y}$. The illustration in \cref{fig:flattness_images} provides a visual representation of this partitioning concept.
Let $\Gamma(\mathbf{y})$ define the local random permutation of pixels within $s\times s$ (\eg, $s=4$) tiles of $\mathbf{y}$. We note that the pixel shuffling is performed only within each tile. We now define the shuffled noisy image $\mathbf{y}^\text{shuffled}$ as:
\begin{equation}
    \mathbf{y}^\text{shuffled} =  \mathbf{c(y)} \odot \Gamma(\mathbf{y}) + (\mathbf{1}- \mathbf{c(y)}) \odot \mathbf{y}
    \label{eq:shuffling}
\end{equation}
Then, the shuffled image can serve as the target for the BSD loss in \cref{eq:loss_bsd}. 
We call this decorrelation technique Local Pixel Shuffling (LPS).
Now we are ready to define the loss of \methodnameacr{} to further reduce the model overfitting caused by highly correlated noise in the BSD approach:

\begin{equation}
    L(\theta) = \mathbb{E}_\mathbf{m} \left[ \| (\mathbf{1} - \mathbf{m}) \odot   (f_{\theta}(\mathbf{y} \odot \mathbf{m}) - \mathbf{y^{\text{shuffled}}}) \|_2^2 \right]
    \label{eq:loss_bsd_reshuffle}
\end{equation}
To determine the region $\Omega_\text{const}$, we adopt a similar approach to \cite{li2023spatially}. We utilize the intermediate pseudo-clean image output $\hat{\mathbf{y}}$, which we obtain from $f_\theta$ (details in the following sections) after a certain number of training iterations. An intuitive indicator of the similarity of pixels within a tile is their standard deviation $\sigma: \Omega \mapsto [0, \infty)$.
Specifically, we calculate the standard deviations within patches of the size of the local tiles independently for each color channel and then average them. 
Finally, we can derive the partition as defined in  \cref{eq:flatness_map}, with the parameter $\lambda>0$ serving as a threshold for the color similarity.
  \begin{equation}
  \Omega_\text{const} = \{ \mathbf{i}:  \sigma[\mathbf{i}] < \lambda\},
    \label{eq:flatness_map}
\end{equation}
\cref{fig:denoising_performance_based_on_shuffling} shows that applying the local pixel shuffling improves the denoising performance when the noise is spatially highly correlated.

\subsubsection{Automated selection of the BSD masking ratio}
\begin{figure*}[t]
    \setkeys{Gin}{width=0.45\linewidth}
    \captionsetup[subfigure]{skip=0.5ex,
                             belowskip=1ex,
                             labelformat=simple}
    \renewcommand\thesubfigure{}

\subfloat[\centering $\beta=0.0$]{\includegraphics[width=0.33\textwidth]{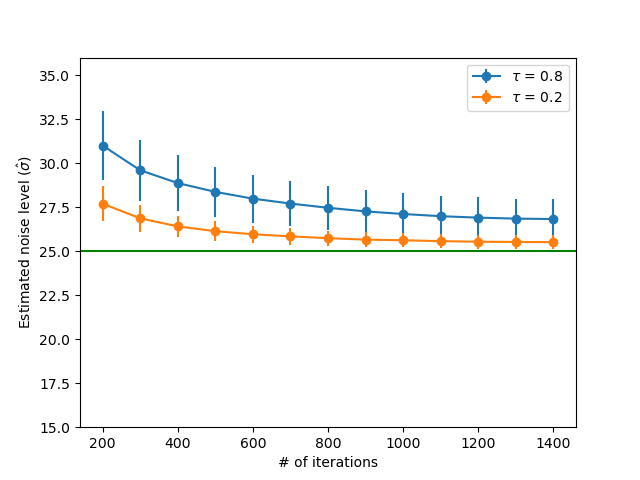}}
\hfill
\subfloat[\centering $\beta=0.5$
]{\includegraphics[width=0.33\textwidth]{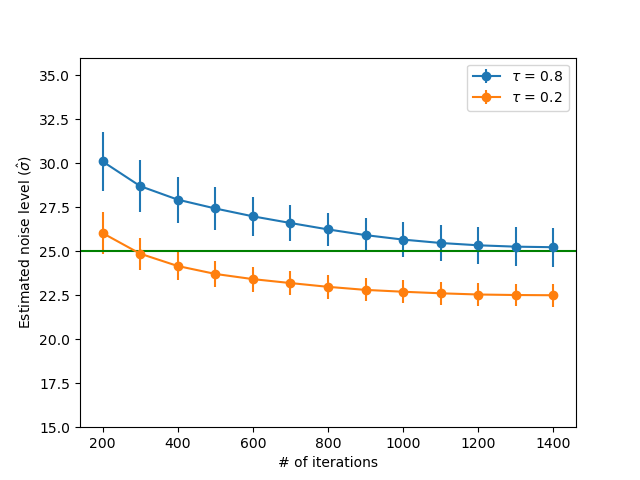}}
\hfill
\subfloat[\centering $\beta=1.0$]{\includegraphics[width=0.33\textwidth]{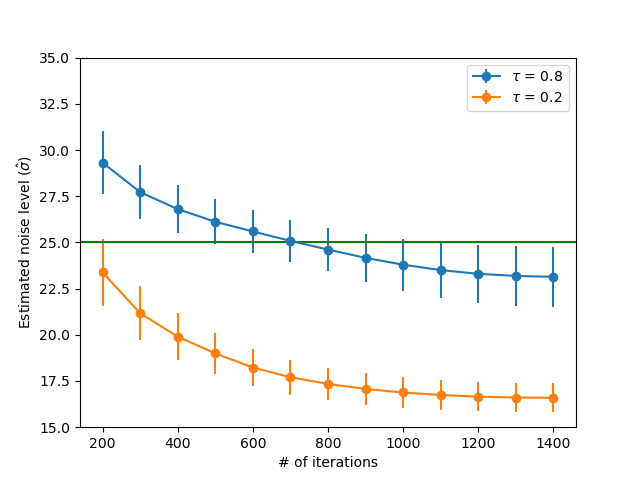}}
\hfill
    \caption{\label{fig:estimated_noise_levels} Estimated noise level based on different correlated noise magnitude and masking ratios.}
    \end{figure*}

To make \methodnameacr{} of practical use, it is essential to have an automated mechanism for determining the noise correlation and, consequently, selecting the optimal masking ratio.
Towards this goal, we analyze the estimated noise level predicted by our BSD method using different masking ratios.
We denote by $\hat\sigma_\tau$ the estimated noise level of the noisy image $\mathbf{y}$ when using a masking scheme with a masking ratio $\tau$:  
\begin{align}
\hat\sigma_\tau = \sqrt{\frac{1}{HWC} \| f_{\theta}(\mathbf{m} \odot \mathbf{y} ) - \mathbf{y}\|_2^2 }.
\end{align}
\cref{fig:estimated_noise_levels} illustrates the estimated noise level $\hat\sigma_\tau$ during the denoising iterations, which is dependent on the estimated parameters $\theta$, for varying correlation regimes and masking ratios. To ensure a dependable indicator, we focus solely on the noise level estimated at convergence and we define the \emph{noise level estimation gap} $\varepsilon$ as the difference in noise levels when utilizing a low masking ratio $\tau^\text{low}$ and a high masking ratio $\tau^\text{high}$:
\begin{align}
\varepsilon = | \hat\sigma_{\tau^\text{high}} - \hat\sigma_{\tau^\text{low}} |
\label{eq:gap}
\end{align}
We set $\tau^\text{low}=0.2$ and $\tau^\text{high}=0.8$.  We experimentally verify that $\varepsilon$ is proportional to the level of correlation in the noise. Therefore, we can use $\varepsilon$ as a proxy for assessing the degree of spatial noise correlation present in the input image $\mathbf{y}$.

\subsection{\methodnameacr{}}

As a conclusion of our prior experimental analysis, we propose to integrate the adaptive masking and local pixel shuffling in the BSD approach. The pseudo-code of our method (\methodnameacr{}) is described in \cref{alg:d2d}.
We begin by training a model $f_\theta$ using two different masking ratios: a high one ($\tau^\text{high}$) and a low one ($\tau^\text{low}$). Subsequently, we calculate the noise level estimation gap $\varepsilon$ as outlined in \cref{eq:gap}. Based on the value of $\varepsilon$, we dynamically determine the masking ratio $\tau$ to utilize, as well as whether to activate the local pixel shuffling. Our automated selection of the optimal masking ratio is depicted in \cref{eq:mask_selection}.
\begin{equation}
\tau^\text{optimal} = 
    \begin{cases}
      \tau^\text{low} & \text{if } \varepsilon \leq \varepsilon^\text{low};\\     
      \tau^\text{medium} & \text{if } \varepsilon^\text{low}<\varepsilon < \varepsilon^\text{high};\\
      \tau^\text{high} & \text{if } \varepsilon^\text{high}<\varepsilon.
    \end{cases}   
     \label{eq:mask_selection}
\end{equation}
If $\varepsilon\le \varepsilon^\text{high}$, we do not apply the local pixel shuffling (indicating low noise correlation) and directly optimize the loss~\eqref{eq:loss_bsd}. If $\varepsilon^\text{high}<\varepsilon$ (implying highly correlated noise), we optimize the loss~\eqref{eq:loss_bsd} for $N_1$ iterations. Subsequently, we determine the partition $\Omega_\text{const}$. We then apply local random permutation $\Gamma$ within $\Omega_\text{const}$ and compute $\mathbf{y}^\text{shuffled}$. Finally, we resume training using the loss~\eqref{eq:loss_bsd_reshuffle}.
The recovered image $\hat{\mathbf{y}}$ is an ensemble of $K$ predictions. To get each prediction, we sample a random binary mask $\mathbf{m_p}$ with a masking ratio $\tau^\text{optimal}$ and apply it to the input image. We set $K=10$.
\begin{equation}
\hat{\mathbf{y}}=\frac{1}{K} \sum_{p=1}^K f_\theta(\mathbf{m_p} \odot \mathbf{y} ). 
\end{equation}

\begin{algorithm}[t!]
 \algsetup{linenosize=\small}
  \scriptsize
  \caption{\methodnameacr{}\label{alg:d2d}}
  \begin{algorithmic}[1]
  \REQUIRE Noisy image $\mathbf{y}$, $f_\theta$,$\tau^\text{high}$, $\tau^\text{medium}$, $\tau^\text{low}$, $\varepsilon^\text{low}$, $\varepsilon^\text{high}$, $N$ and $N_1<N$
  \ENSURE Restored image $\hat{\mathbf{y}}$\\
 \STATE Apply the BSD baseline with $\tau^\text{high}$ and $\tau^\text{low}$ respectively.
  \STATE Compute  $\varepsilon$ from eq.~\eqref{eq:gap}
 \STATE   $\tau^\text{optimal} = $
    $\begin{cases}
      \tau^\text{low} & \text{if } \varepsilon \leq \varepsilon^\text{low};\\     
      \tau^\text{medium} & \text{if } \varepsilon^\text{low}<\varepsilon < \varepsilon^\text{high};\\
      \tau^\text{high} & \text{if } \varepsilon^\text{high}<\varepsilon.
    \end{cases} $ 
  
 
  \IF{ $\varepsilon< \varepsilon^\text{high} $}
 \FOR{$t: 1 \rightarrow N$} 
  \STATE Update $f_\theta$ by optimizing eq.~\eqref{eq:loss_bsd} using $\tau^\text{optimal}$
  
  \ENDFOR\\
 \ELSE
   \FOR{$t: 1 \rightarrow N_1$} 
  \STATE Update $f_\theta$ by optimizing eq.~\eqref{eq:loss_bsd} using $\tau^\text{optimal}$
   \ENDFOR\\
  \STATE compute the partition $\Omega_\text{const}$ using eq.~\eqref{eq:flatness_map}
  \STATE compute $\mathbf{y}^\text{shuffled}$ using eq.~\eqref{eq:shuffling}
  \FOR{$t: N_1 \rightarrow N$} 
   \STATE Update $f_\theta$ by optimizing eq.~\eqref{eq:loss_bsd_reshuffle} using $\tau^\text{optimal}$
  \ENDFOR\\
  \ENDIF

  \STATE Return the restored image $\hat{\mathbf{y}}=\frac{1}{K} \sum_{p=1}^K f_\theta( \mathbf{m_p} \odot \mathbf{y})$\\
\end{algorithmic}

\end{algorithm}

\section{Experiments}
In this section, we will first introduce the experimental settings. We will then present the quantitative and qualitative results of \methodnameacr{}, along with comparisons with other methods.
\subsection{Experimental settings}
\noindent\textbf{Datasets} We evaluated our method on four widely-used real-world noise datasets: SIDD (validation and benchmark datasets) \cite{SIDD_2018_CVPR}, FMDD \cite{zhang2019poisson}, and PolyU \cite{nam2016holistic}. The validation and benchmark datasets of SIDD contain natural sRGB images captured by smartphones, with each dataset consisting of 1280 patches sized $3\times 256\times 256$. FMDD contains fluorescence microscopy images with a size of $512\times512$. The PolyU dataset consists of 100 natural images taken from diverse commercial camera brands, with each image having a size of $3\times 512\times 512$.

\noindent\textbf{Implementation details}
The network architecture for \methodnameacr{} is the same as in Noise2Noise \cite{lehtinen2018n2n}. The denoising network is trained from scratch using the Adam optimizer with cosine annealing. By default, we use the following hyperparameters in our implementation unless otherwise specified: $\tau^\text{high} = 0.8$, $\tau^\text{low} = 0.2$, $\tau^\text{medium} = 0.5$, $\varepsilon^\text{low} = 1.5$, $\varepsilon^\text{high} = 2.5$, $s=4$ and $N = 800$. In our supplementary material, we provide a more in-depth discussion on the selection of hyperparameters.

\begin{figure}[t]
    \setkeys{Gin}{width=0.33\linewidth}
    \captionsetup[subfigure]{skip=0.5ex,
                             belowskip=0.5ex,
                             labelformat=simple}
    \renewcommand\thesubfigure{}
    \setlength\tabcolsep{1.pt}
    \small
    \centering
  \begin{tabular}{ccccc}

 {\includegraphics{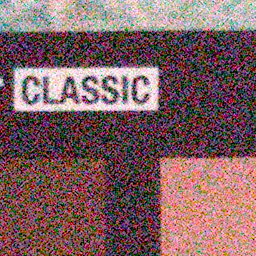}}&  {\includegraphics{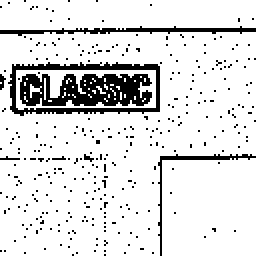}} & %
{\includegraphics{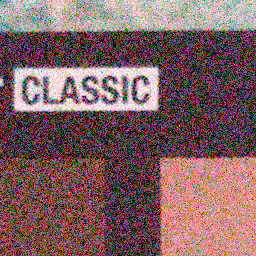}}\\


{\includegraphics{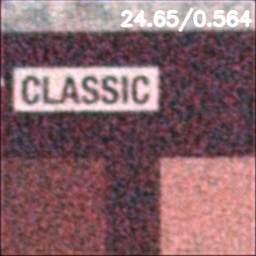}}& {\includegraphics{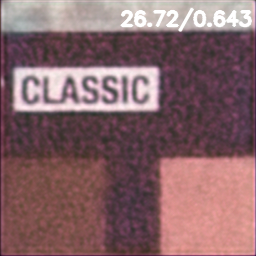}}& {\includegraphics{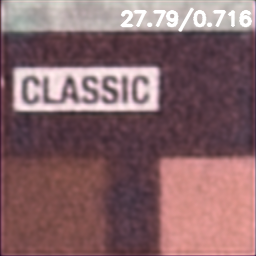}}\\

   \end{tabular}
\captionof{figure}{\label{fig:real_world1} Top: from left to right: original noisy image, Mask capturing region flatness derived from pseudo-clean, Shuffled noisy image (using LPS). Bottom: from left to right: Result using the baseline ($\tau=0.5$), Result without local pixel shuffling , Ours.}
\end{figure}

\begin{table*}[h!]
  \begin{center}
    \renewcommand{\arraystretch}{1.2}
     \begin{tabularx}{\textwidth} {@{\hspace{2.em}}c@{\hspace{1.em}}|c@{\hspace{1.em}}c@{\hspace{1.em}}c@{\hspace{1.em}}c@{\hspace{1.em}}c@{\hspace{2.em}}}

    \cline{1-6}
    {Category} & {Method} & {SIDD Validation}& {SIDD Benchmarck}& {FMDD}& {PolyU}\\
     \cline{1-6}
     
    &  BM3D \cite{dabov2009bm3d} & {25.65/0.475}  & {25.65/0.685} & {30.06/0.771} & { 37.40/0.953}  \\ 
     
      &  DIP \cite{ulyanov2018dip} & { 32.11/0.740} & {-} & {32.90/0.854} & {37.17/0.912}  \  \\ 
      
    {  Single Image} & Self2Self \cite{quan2020self2self} & {29.46/0.595} & {29.51/0.651} & {30.76/0.695} & {37.52/0.926}   \\
   (test-time training) &  PD-denoising \cite{zhou2020awgn} & {33.97/0.820} & {33.61/0.894} & {33.01/0.856} & {37.04/0.940}  \\
    &  NN+denoiser \cite{zheng2020unsupervised} & {-} & {33.18/0.895} & {32.21/0.831} & {\underline{37.66}/\underline{0.956}}  \\
    &  APBSN-single \cite{lee2022ap} & { 30.90/0.818} & { 30.71/0.869} & {28.43/0.804} & {29.61/0.897}  \\
     & ScoreDVI \cite{cheng2023score} & { \underline{34.75}/\textbf{0.856}} & { 34.60/0.920} & { \underline{33.10}/\underline{0.865}} & {\textbf{37.77}/\textbf{0.959}}  \\   
    &  Baseline & { 33.12/0.805} & { 32.67/0.850} & {32.25/0.824} & {37.12/0.911}  \\
     &  Ours & { \textbf{35.06}/\underline{0.851}} & { \textbf{34.78}/\underline{0.900}} & {\textbf{33.71}/\textbf{0.882}} & {37.62/0.932}  \\
    \cline{1-6}

      {Noisy/Impaired} &  APBSN \cite{lee2022ap} & {-} & \textbf{36.91/0.931} & \underline{31.99/0.836} & \textbf{37.03/0.951} \\ 
      {Dataset} &  CVF-SID \cite{neshatavar2022cvf} & \underline{34.81/0.944} & {34.71/0.917} & \textbf{32.73/0.843} & {35.86/0.937}   \\
        & LUD-VAE \cite{zheng2022learn} & { \textbf{34.91/0.944} } & \underline{34.82/0.926} & {-} & \underline{36.99/0.955}   \\
       \cline{1-6}
     {Supervised} &  DnCNN \cite{thakur2019state} & {37.73 / 0.943} & {37.61 / 0.941} & {-} & {-}   \\\cline{1-6}

    \end{tabularx}
    \caption{\label{tab:real-results}Quantitative comparisons (PSNR(dB)/SSIM) of our method and other real-world denoising methods including single image-based methods and dataset-based methods on SIDD, FMDD, PolyU datasets. The best results of the unsupervised approaches are marked in \textbf{bold}, while the second best ones are \underline{underlined}.}

  \end{center}

\end{table*}

\begin{figure*}[h!]
    \setkeys{Gin}{width=0.135\linewidth}
    \captionsetup[subfigure]{skip=0.01ex,
                             belowskip=0.5ex,
                             labelformat=simple}
    \renewcommand\thesubfigure{}
    \setlength\tabcolsep{1.5pt}

\small
  \begin{tabular}{ccccccc}

  \textit{Noisy}  & \textit{ DIP~\cite{ulyanov2018dip} } & \textit{Self2Self~\cite{quan2020self2self}} & \textit{NN+denoiser~\cite{zheng2020unsupervised}} & \textit{Baseline} & \textit{Ours} & \textit{Ground-truth }\\
  {\includegraphics{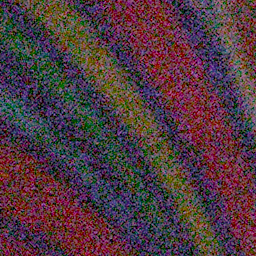}} &
{\includegraphics{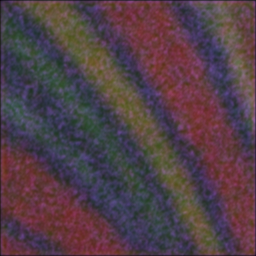}}& {\includegraphics{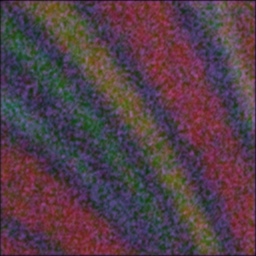}}& {\includegraphics{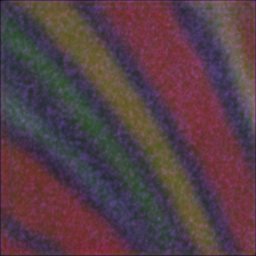}}& {\includegraphics{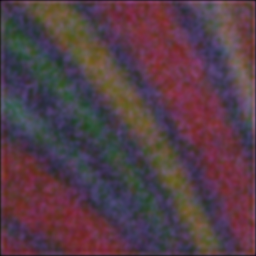}}& {\includegraphics{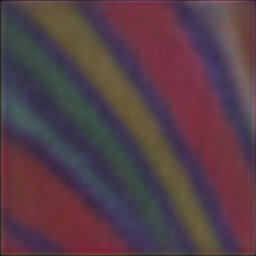}} &
{\includegraphics{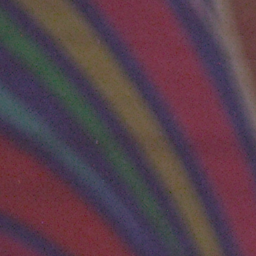}}  \\
\textit{} &  \textit{26.64/0.546} & \textit{25.78/0.579} & \textit{28.35/0.646} & \textit{28.13/0.663} & \textit{\textbf{31.79}/\textbf{0.812}} & {} \\


 {\includegraphics{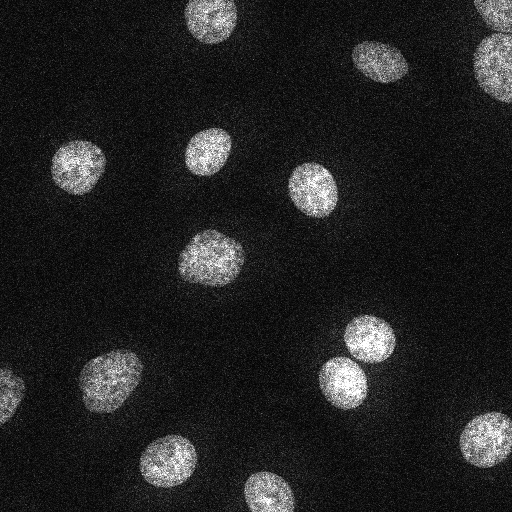}} & {\includegraphics{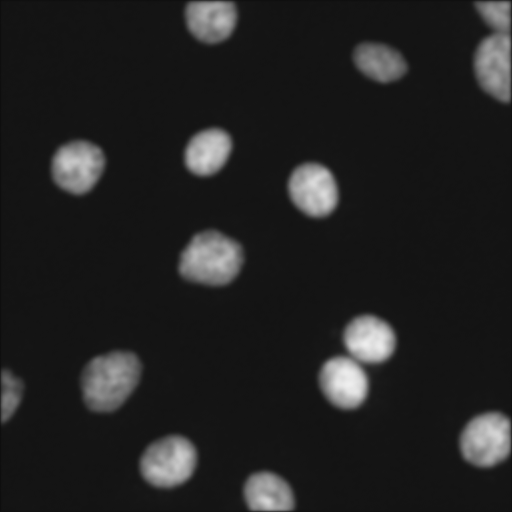}}& 
{\includegraphics{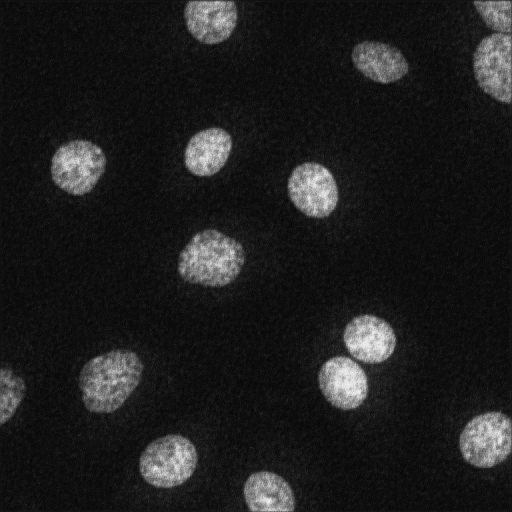}}&
{\includegraphics{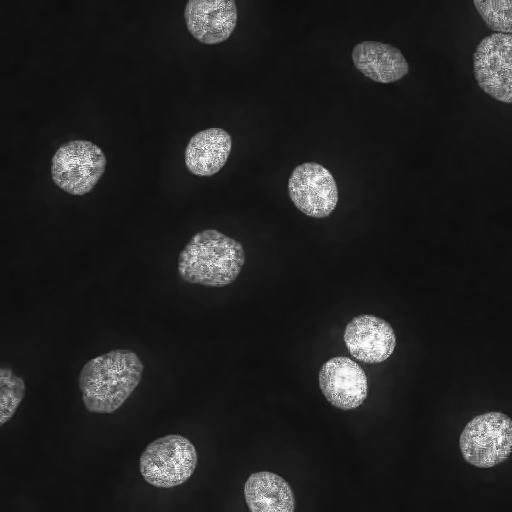}}& 
{\includegraphics{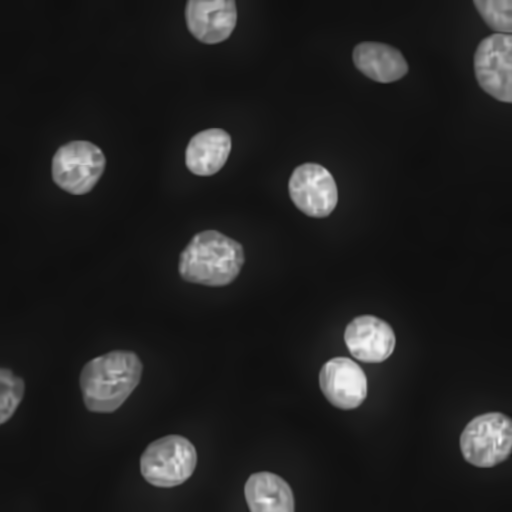}}& {\includegraphics{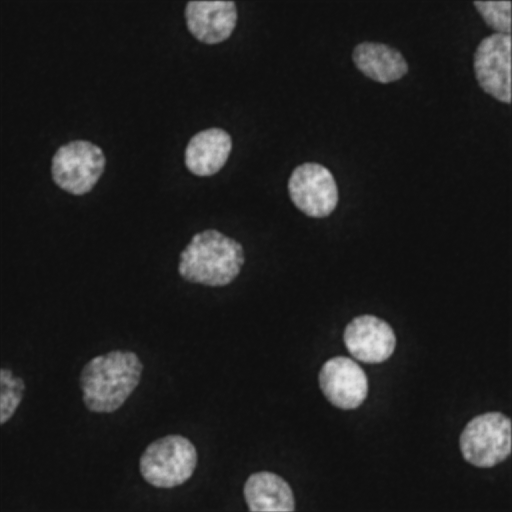}} &
{\includegraphics{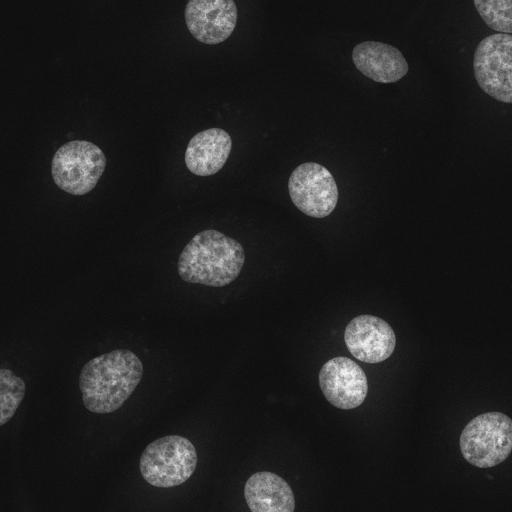}} \\
 & {29.67/0.908} & \textit{28.79/0.854} & \textit{30.68/0.921} & \textit{30.13/0.918} & \textit{\textbf{32.05}/\textbf{0.937}} &  \\

 {\includegraphics{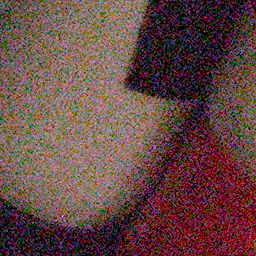}} &
{\includegraphics{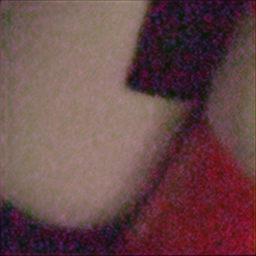}}& {\includegraphics{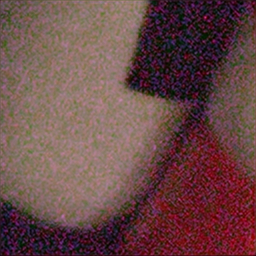}}& {\includegraphics{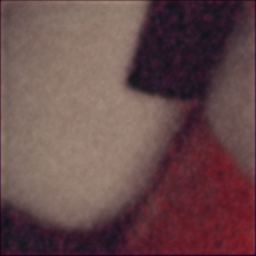}}& {\includegraphics{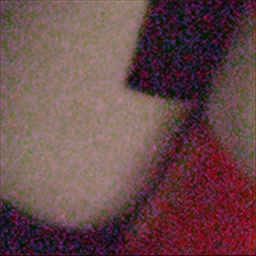}}& {\includegraphics{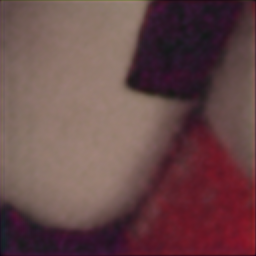}} &
{\includegraphics{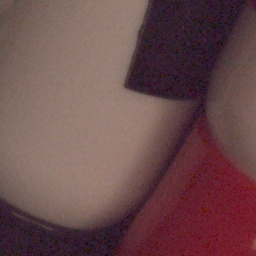}}  \\
 & {28.04/0.622} & \textit{24.36/0.410} & \textit{30.14/0.751} & \textit{25.34/0.467} & \textit{\textbf{31.42}/\textbf{0.793}} &  \\

   \end{tabular}

\captionof{figure}{\label{fig:real_noise_results} Visual comparison of our method against other single image-based denoising methods in SIDD validation and FMDD datasets.
The PSNR/SSIM results are reported under each image.}
\end{figure*}

\subsection{Evaluation on real-world noise}

 In our evaluation, we compare our method against several single image-based denoising methods, including Self2Self \cite{quan2020self2self}, NN+denoiser \cite{zheng2020unsupervised}, DIP \cite{ulyanov2018dip}, BM3D \cite{dabov2009bm3d}, PD-denoising \cite{zhou2020awgn}, NN+denoiser \cite{zheng2020unsupervised}, scoreDVI \cite{cheng2023score}, and APBSN-single \cite{lee2022ap}. For APBSN-single, we adapt the APBSN method from \cite{lee2022ap} to directly denoise a single image. The strides of PD in training and testing are 5 and 2, respectively. For NN+denoiser \cite{zheng2020unsupervised}, we use its best version, which is NN+BM3D for single image denoising. 
 For the other methods, we either use the authors' code or directly adopt their published results if available. We also include a comparison with a baseline method (denoted as Baseline in Table) which is a blind spot method with the same network architecture as our method but with a fixed masking ratio of $\tau=0.5$. 
Additionally, we compare against dataset-based methods including CVF-SID \cite{neshatavar2022cvf}, LUD-VAE \cite{zheng2022learn}, and APBSN \cite{lee2022ap}. We also provide a reference comparison with the supervised DNCNN \cite{thakur2019state}. 
 The quantitative comparisons are summarized in \cref{tab:real-results}. Qualitative comparisons among different single image-based methods on SIDD and FMDD are displayed in \cref{fig:real_noise_results}. Additional visual comparisons are included in the supplementary material.
\methodnameacr{} shows a significant boost over the baseline, with an improvement of about 2 dB for both SIDD datasets and 1.5 dB for the FMDD dataset, highlighting the importance of our adaptive masking scheme and local pixel shuffling. \methodnameacr{} yields competitive results compared to existing single image-based methods. Our method excels in the FMDD dataset by outperforming both the single-image and dataset-based methods, which encompass images with varying noise levels and correlations. 
\cref{fig:real_world1} shows the output of \methodnameacr{} on an image from the SIDD validation dataset. Utilizing our adaptive masking scheme and pixel local shuffling results in a significant improvement over the baseline.
\begin{table}[t]
      \centering
    \small
\begin{tabular}{@{}lcc@{}}
\toprule
     & SIDD & FMDD  \\ \midrule
Adaptive masking accuracy      &    88.7 \%        &   92.4 \%         \\
\bottomrule
\end{tabular}

\caption{Adaptive masking accuracy on SIDD and FMDD datasets. \label{tab:masking_accuracy}}
\end{table}
\begin{table}[ht]
   
    \centering
    \small
   \begin{tabular}{@{}ccccc@{}}
\toprule
Adaptive masking ratio    & Local pixel shuffling & SIDD & FMDD \\ \midrule
No      &       No     &   33.12     &  32.25    &   \\
No      &    Yes   &  33.86    &  32.92    & \\
Yes     &       No    &   34.45     &  33.56     &   \\
Yes    &     Yes      &  35.06     &   33.71     &   \\
\bottomrule
\end{tabular}
\caption{Ablation of \methodnameacr{} components .\label{tab:ablation1}}
    
\end{table}

\section{Ablations}
We perform ablation studies to analyze the impact of each component of our method.
\subsection{Influence of Adaptive Masking}
We evaluate the effectiveness of our adaptive masking scheme. When the adaptive masking is not applied, we utilize a default masking ratio of $\tau=0.5$. \cref{tab:ablation1} demonstrates that adaptive masking yields a notable performance enhancement for both the SIDD and FMDD datasets. 
We further analyze the effects on each image individually by calculating the success rate of our adaptive masking scheme. To do this, we denoise each image using the baseline with three masking ratios: $\tau^\text{high} = 0.8$, $\tau^\text{medium} = 0.5$, and $\tau^\text{low} = 0.2$, and we store the best result. Then, we denoise each image using \methodnameacr{}. In \cref{tab:masking_accuracy}, we show how many times our algorithm succeeds in selecting the optimal masking ratio that leads to the best performance. Our adaptive masking is considered successful if it corresponds to the masking ratio that results in optimal performance. \cref{tab:masking_accuracy} indicates that our adaptive masking succeeds in selecting the optimal masking ratio in approximately 90\% of cases. Further analysis on the robustness of \methodnameacr{} hyperparameters and showcasing of failure cases can be found in our Supplementary materials.

\subsection{Influence of Local Pixel Shuffling}
We validate the significance of local pixel shuffling in our method. As shown in \cref{tab:ablation1}, there is an improvement of approximately 0.7 dB for both the SIDD and FMDD datasets when applying local pixel shuffling.

\subsection{Influence of the neighberhood size $s$}
\cref{tab:ablation_neighberhood size} shows an ablation of the neighborhood size $s$ conducted on the SIDD valadition dataset. As $s$ increases, the performance improves then it starts to saturate. We note that $s$ is an hyperparameter related to the local pixels shuffling which is applied only if a high spatial correlated noise is detected.
\begin{table}[t]
      \centering
    \small
   \begin{tabular}{@{}lccccc@{}}
\toprule
  & $ s=2$ & $s=3$ & $s=4$& $s=5$  & $s=6$ \\ \midrule
PSNR      &    34.65         &   34.85      &   35.06 &   35.15 &   35.09 \\
\bottomrule
\end{tabular}

\caption{Ablation of the neighberhood size $s$ on the SIDD validation dataset. \label{tab:ablation_neighberhood size}}
\end{table}

\subsection{Computational Efficiency}
We provide an analysis of the inference time, model parameters and floating-point operations per second (FLOPs) of the compared deep learning-based methods. According to the results in \cref{tab:real-results} and \cref{tab:efficiency}, our method demonstrates a balance between effectiveness and efficiency. It is worth noting that there is room for speed enhancement with \methodnameacr{}. For example, the initial training phase to determine the optimal masking could be performed in parallel rather than sequentially.

\begin{table}[t]
\small
    \centering
   \begin{tabular}{@{}lcccc@{}}
\toprule
Method     & Infer. time (s) & Params (M) & FLOPs (G) \\ \midrule
DIP~\cite{ulyanov2018dip}       &       146.2     &   13.4     &  31.06    &   \\
Self2Self~\cite{quan2020self2self}       &     3546.5   &  1.0     &  9.55     & \\
NN+denoiser~\cite{zheng2020unsupervised}      &        897.6    &   13.4     &  31.06     &   \\
APBSN-single~\cite{lee2022ap}   &     121.4      &  3.66     &   234.63     &   \\
ScoreDVI~\cite{cheng2023score}   &     81.2      &  13.5     &  37.87     &   \\
Baseline      &      24.6      &     0.99     &  11.44\\
Ours      &      75.3      &     0.99     &  11.44     & \\ \bottomrule
\end{tabular}
\caption{Efficiency comparisons of deep learning-based
methods under the input size 256 × 256 × 3.\label{tab:efficiency}}

\end{table}

\section{Conclusion}
We have introduced \methodnameacr{}, a single image denoising method that leverages the blind spot denoising framework. Our approach includes an analysis to detect and alleviate the impact of noise correlation. We demonstrated that the masking ratio plays a critical role in denoising performance, especially in the presence of correlated noise. Building on this analysis, we proposed a method to automatically estimate the level of noise correlation. Additionally, we introduced a technique to directly de-correlate noise in the input image by shuffling pixels with similar denoised color intensities. As a result, our method, \methodnameacr{}, achieves state-of-the-art results compared to existing test-time training approaches across various public benchmarks.

\noindent\textbf{Acknowledgements} We acknowledge the support of the SNF project number 200020\_200304.
{
    \small
    \bibliographystyle{ieeenat_fullname}
    \bibliography{main}
}


\end{document}